\pdfoutput=1

\documentclass[11pt]{article}

\usepackage[final]{acl}

\usepackage{times}
\usepackage{latexsym}

\usepackage[T1]{fontenc}

\usepackage[utf8]{inputenc}

\usepackage{microtype}

\usepackage{inconsolata}

\usepackage{graphicx}
\usepackage{epstopdf} 

\usepackage{float}
\usepackage{stfloats}
\usepackage{array}
\usepackage{svg}
\usepackage{xcolor}
\usepackage{colortbl}
\usepackage{todonotes}

\title{Teaching Large Language Models Number-Focused Headline Generation With Key Element Rationales}

\author{Zhen Qian \and Xiuzhen Zhang\thanks{Corresponding author} \and Xiaofei Xu \and Feng Xia \\
         School of Computing Technologies, RMIT University, Australia \\ 
         \texttt{\{s3888611,xiaofei.xu2\}@student.rmit.edu.au;} \\
         \texttt{\{xiuzhen.zhang,feng.xia\}@rmit.edu.au}
}

\begin{document}
\maketitle
\begin{abstract}
Number-focused headline generation is a summarization task 
that requires both high textual quality and precise numerical accuracy, which poses a unique challenge for Large Language Models (LLMs). 
Existing studies in the literature focus only on either textual quality or numerical reasoning and thus are inadequate to address this challenge. 
In this paper, we propose a novel chain-of-thought framework for using rationales comprising key elements of the Topic, Entities, and Numerical reasoning (TEN) in news articles to enhance the capability  
for LLMs to generate topic-aligned high-quality texts with precise numerical accuracy. 
Specifically, a teacher LLM is employed to generate TEN rationales as supervision data, which are then used to teach and fine-tune a student LLM. 
Our approach teaches the student LLM automatic generation of rationales with enhanced capability for numerical reasoning and topic-aligned numerical headline generation.  
Experiments show that our approach achieves superior performance 
in both textual quality and numerical accuracy.  
Our implementation is publicly available at 
\texttt{https://github.com/TEN-Sum/TEN}.
\end{abstract}

\section{Introduction}

Headline generation, an important task in abstractive summarization, aims to condense a news article into a single line of text. 
In the literature, text summarization models employ pre-trained language models \citep{lewis_bart_2019, raffel_exploring_2023, zhang_pegasus_2020} and large language models (LLMs)~\cite{jin_comprehensive_2024} have shown 
high textual quality for headline generation. 

Numerical facts are crucial elements for modern news articles, and headlines often include numerals to enhance conciseness and attract readers' attention. A headline like "Pink Floyd reaches deal with Sony to sell music catalogue for \underline{\$400m}"\footnote{https://www.theguardian.com/music/2024/oct/02/pink-floyd-catalog-sony} immediately grabs readers' interest. 

\begin{figure}[t]
  \centering
  \includegraphics[width=\columnwidth]{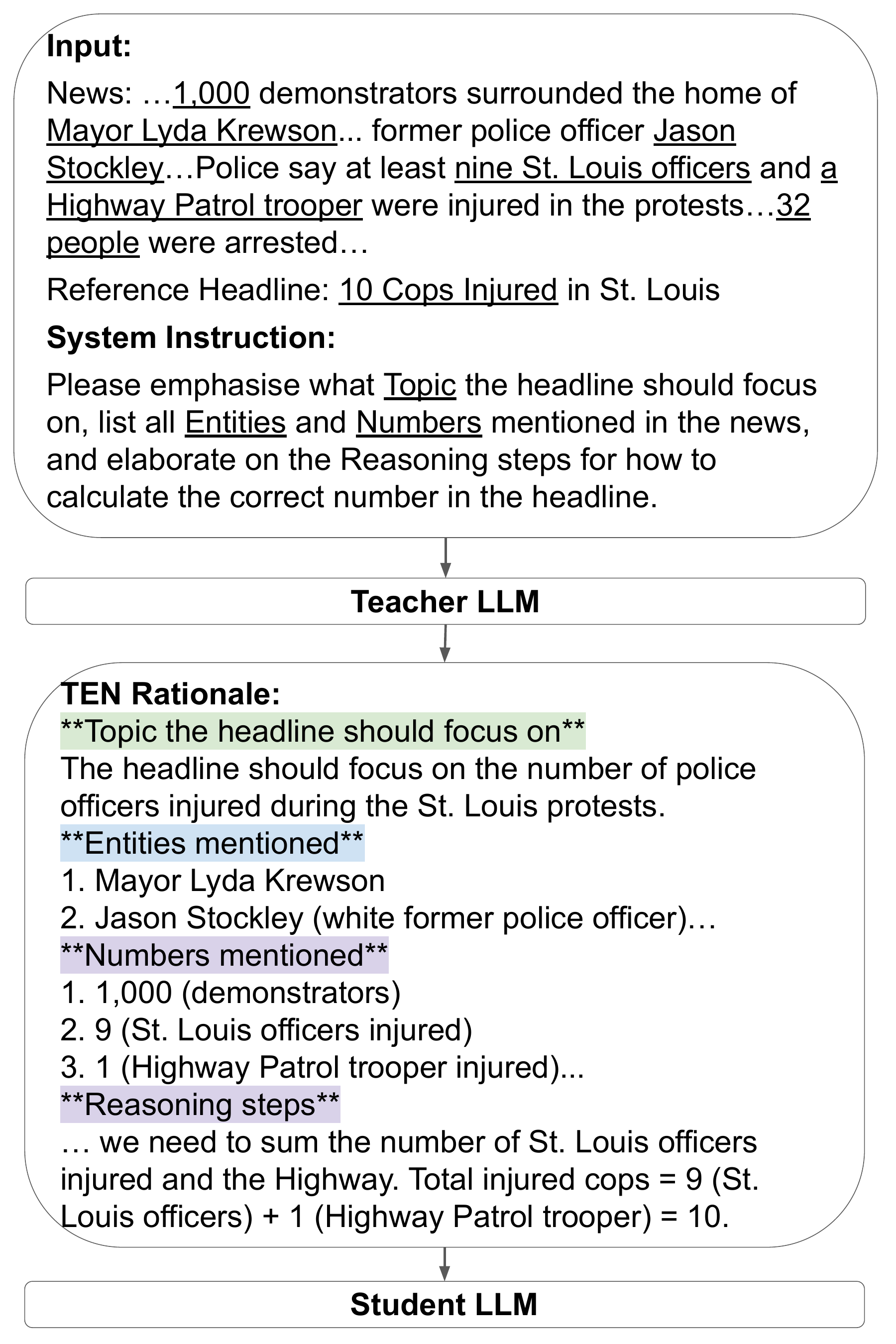}
 \caption{An example TEN Rationale for key elements of \underline{T}opic (green), \underline{E}ntities (blue) and \underline{N}umerical reasoning (purple).}
  \label{f1}
\end{figure}

Research shows that generating headlines with correct numbers requires mathematical reasoning capabilities in text summarization models~\cite{huang-etal-2024-numhg}. 
Obtaining correct numbers in headlines can involve mathematical operations such as addition, subtraction, and rounding of numbers from the source news articles. 
As shown in Fig.~\ref{f1}, the news article covers the St. Louis protests, mentioning various entities and numbers. To generate an accurate headline, the language model must first identify the most newsworthy aspect of the event—the number of injured police officers. It then needs to calculate the correct number based on the information provided. In this case, the headline's number, 10, is not explicitly stated but requires addition (9 plus 1).
Number-focused headline generation requires 
not only text summarization to produce high quality text but also numerical reasoning within the textual context to generate the correct numbers. 

Existing studies on text summarization and numerical reasoning are inadequate for this challenging numerical headline generation problem~\citep{huang-etal-2024-numhg}. For text summarization, state-of-the-art pre-trained language models (PLMs)~\citep{lewis_bart_2019,zhang_pegasus_2020,liu_brio_2022} have relied on supervised fine-tuning to develop their summarization abilities. Researchers have also applied Chain-of-Thought (CoT) prompting~\citep{wang_element-aware_2023}, reinforcement learning~\citep{stiennon_learning_2022}, and direct preference optimization (DPO)~\citep{rajpoot_team_2024} to large language models (LLMs), aiming to improve their summarization quality. However, these methods focus on textual quality and do not address numerical accuracy.

On the other hand, numerical reasoning models mainly focus on tasks that require producing a single numerical answer, such as solving word math problems, rather than generating text that includes numbers~\citep{ling_program_2017,amini_mathqa_2019,chiang_semantically-aligned_2019, cobbe_training_2021,wei_chain--thought_2023}. 
Researchers have shown that language models' proficiency in solving these tasks can be enhanced through 
explanation of intermediate steps
~\citep{amini_mathqa_2019, chiang_semantically-aligned_2019, cobbe_training_2021, wang_t-sciq_2023}, verification~\citep{cobbe_training_2021, wang_math-shepherd_2024}, and reinforcement learning~\citep{wang_math-shepherd_2024}. 
However, these techniques are developed in a setting where the question is given and they only need to infer the correct number as the final output. 


In this paper, we propose a novel Chain-of-Thought (CoT) framework that uses rationales comprising key elements of \underline{T}opic, \underline{E}ntity and \underline{N}umerical reasoning (TEN) to teach and fine-tune LLMs for number-focused headline generation. 
Here rationales refer to textual descriptions for the key elements in a news article -- topics, entities, and numbers and their intermediate reasoning steps. These key element rationales can be used to enhance LLMs for the generation of topic-aligned headlines with higher numerical accuracy.  

Instead of costly manual annotation of TEN rationales, we propose to fine-tune open-source LLMs (e.g. Mistral 7B) to automatically generate such rationales for numerical headline generation.   
To enhance the capability for an open-source LLM to generate TEN rationales, we employ the teacher-student knowledge distillation framework~\cite{wang_t-sciq_2023} and leverage a powerful teacher LLM (e.g. GPT 4o) to generate TEN rationals as supervision data to fine-tune the open-source LLM as a student. 
Experiments show that our approach can achieve significant improvement over strong baselines in both textual quality and numerical accuracy. 


Contributions of our research are three fold:
\begin{itemize}
    \item We propose a CoT framework that uses rationales of key elements Topic, Entities and Numerical reasoning for LLMs to generate number-focused headlines. 
    \item To enhance the capability for LLMs to generate topic-aligned headline text with high numerical accuracy, we apply the teacher-student framework to distill knowledge from a powerful teacher LLM to fine-tune an open-source LLM for automatic generation of TEN rationales.   
    \item We further develop a strategy based on Direct Preference Optimization \citep{rafailov_direct_2023} for LLMs to refine generation of TEN rationales. 
\end{itemize}

\section{Related Work}

\subsection{Headline Generation}

Headline generation, a form of extreme text summarization, requires producing highly condensed, single-sentence summaries that capture the key information in a news article. 
In early studies~\citep{rush_neural_2015,narayan_dont_2018}, models are supervise-trained on datasets containing single-sentence summaries (XSum)~\citep{narayan_dont_2018} and news-headline pairs (Gigaword) \citep{rush_neural_2015} However, their CNN-based and RNN-based approaches have since been outperformed by transformer-based models. 
Recent studies show that transformer-based PLMs such as BART~\citep{lewis_bart_2019}, PEGASUS~\citep{zhang_pegasus_2020}, and BRIO~\citep{liu_brio_2022} can be fine-tuned on XSum and Gigaword to achieve promising results for extreme summarization and headline generation. 

While PLMs laid the bedrock for summarization, the advancement of LLMs has pushed the boundaries further. Several LLM-based approaches have emerged for general text summarization. Recent works have leveraged CoT prompting for summarization, proposing a "Summary Chain-of-Thought" method that guides LLMs to focus on key elements and generate summaries step-by-step~\citep{wang_element-aware_2023}. To further enhance summary quality, reinforcement learning methods have been employed to optimize LLMs based on human preferences~\citep{stiennon_learning_2022}. LLM-based approaches have also been tailored specifically for headline generation. For instance, leveraging reinforcement learning, \citet{tan_enhancing_2024} focus on creating personalized headlines for content recommendation. 
These approaches, whether PLM-based or LLM-based, focus on the text quality of summarization and the numerical accuracy is overlooked.

Research on number-focused headline generation is reported recently. 
\citet{huang-etal-2024-numhg} assess the performance of PLMs in number-focused headline generation, but they do not provide strategies to enhance the models' numerical accuracy. \citet{rajpoot_team_2024} apply DPO to optimize headline generation using a preference dataset designed to train the model to favor headlines with correct numbers. While this preference for correct numbers can improve numerical accuracy, solely relying on it may degrade the textual quality of the generation. 
\subsection{CoT Prompting for Rationale Generation}

CoT prompting has gained great popularity due to its potential to unlock LLMs' reasoning capabilities by simply instructing them to generate intermediate steps as rationales before reaching a final answer~\citep{wei_chain--thought_2023}. For example, one can utilize CoT reasoning by simply adding the phrase "let's think step by step" to the end of each question~\citep{kojima_large_2023}. This approach is improved by a two-step process to generate rationales~\citep{zhang_automatic_2022}: first, selecting representative questions to generate exemplar rationales, and then using these representative rationales as demonstrations for LLMs to generate reasoning steps for other questions in the dataset. This idea is further enhanced by including the correct solution in prompts can enhance the quality of LLM-generated rationales~\citep{magister_teaching_2023}.



Especially for word math problems, research shows that LLM's numerical reasoning ability can be improved by learning from human-crafted rationales, including natural language intermediate reasoning steps~\citep{ling_program_2017} and symbolic representations like equations~\citep{chiang_semantically-aligned_2019,amini_mathqa_2019}. 
But these methods rely on human annotations and therefore are costly. 

\subsection{Learning from Teacher LLM Generated Rationales}

Learning from rationales generated by teacher LLMs is a scalable alternative to human annotation. 
Research has shown that CoT reasoning steps generated from teacher LLMs can be used to fine-tune smaller student language models~\cite{ho_large_2023,hsieh_distilling_2023} that 
may even outperform the teacher LLM for some tasks.
Such teacher-student knowledge distillation has also been applied to multi-modal training for science QA~\cite{wang_t-sciq_2023}, which involves numerical reasoning.
The authors propose that mixing simple and complex reasoning in supervision data can enhance student LLMs' performance~\cite{wang_t-sciq_2023}. 
Our approach also leverages the teacher-student knowledge distillation framework. 
Unlike existing work, we focus on rationales for topic alignment as well as numerical reasoning in numerical text generation.

Researchers have explored various approaches to enhance rationale quality, including the use of verifiers~\cite{cobbe_training_2021}, majority voting~\cite{wang_self-consistency_2023} and reinforcement learning~\cite{wang_math-shepherd_2024}. 
Our approach, which leverages DPO for refining rationale generation, is closely related to reinforcement learning strategies~\citep{wang_math-shepherd_2024}. 
However, unlike previous work that focuses solely on reward models for numerical reasoning rationales, our approach develops preference datasets tailored to both nuanced topic alignment and complex numerical reasoning. 

\begin{figure*}[t]
  \centering
  \includegraphics[width=\textwidth, trim={2cm 2.45cm 0cm 1.7cm}, clip]{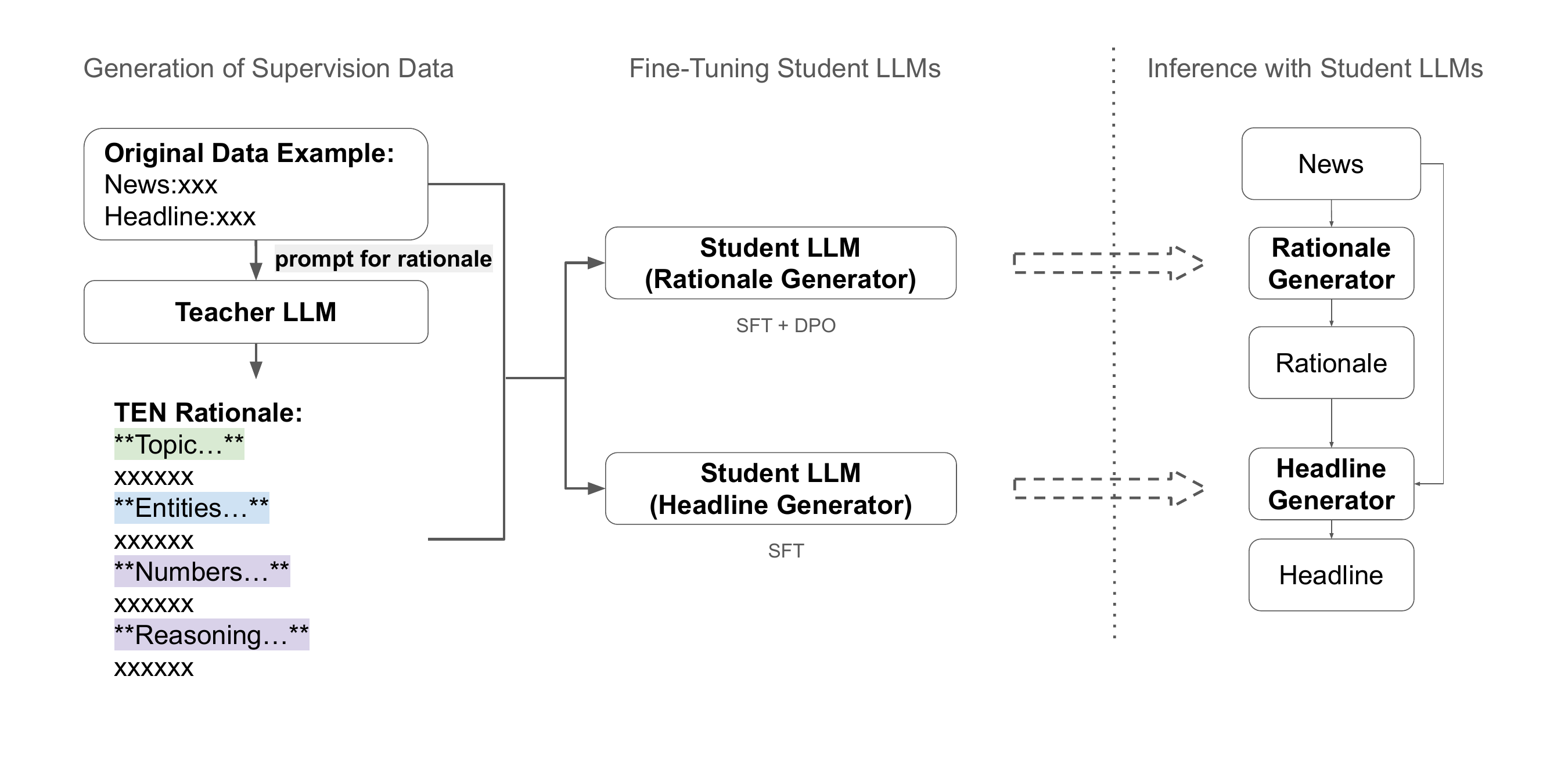}
  \caption {Our TEN approach for automatic generation of rationales to enhance numerical headline generation.}
  \label{fig:three_phase_of_TEN}
\end{figure*}

\section{Methodology}

This section presents our framework for leveraging the teacher-student knowledge distillation framework to fine-tune LLMs for automatic generation of TEN rationales to enhance LLM headline generation.
We employ a teacher LLM (e.g. GPT 4o) to generate TEN rationales 
and use these rationales as supervision data to fine-tune a student LLM (e.g. Mistral-7B), including a rationale generator for automatic generation of TEN rationales and
a headline generator for headline generation. 

As shown in Fig.~\ref{fig:three_phase_of_TEN}, our approach adopts a teacher-student framework to fine-tune a (student) LLM to automatically generate TEN rationales.  
When generating supervision data to fine-tune a rationale generator, 
we prompt a teacher LLM to generate rationales for each news-headline pair in the dataset.  
The rationales are aimed to enhance the topic alignment and numerical reasoning capabilities for numerical headline generation, comprising key elements \underline{T}opic and \underline{E}ntities, 
as well as \underline{N}umbers in the news article and the intermediate reasoning steps to calculate the correct number in the headline. 

The teacher LLM generated rationales are used as supervision data to 
fine-tune a student LLM as the rationale generator. 
We further refine the rationale generator using DPO. The preference data for DPO are automatically generated to favour rationales that lead to headlines with matching topics and accurate numerals. 
The news article and teacher LLM generated TEN rationales are then used to fine-tune another student LLM for headline generation. 
In the inference phase, the two fine-tuned student LLMs are used sequentially. The rationale generator will first produce TEN rationales for the input news article. The headline generator will then use the rationales together with the news articles as input to generate final headlines. 

\begin{figure*}[t]
  \includegraphics[width=\textwidth]{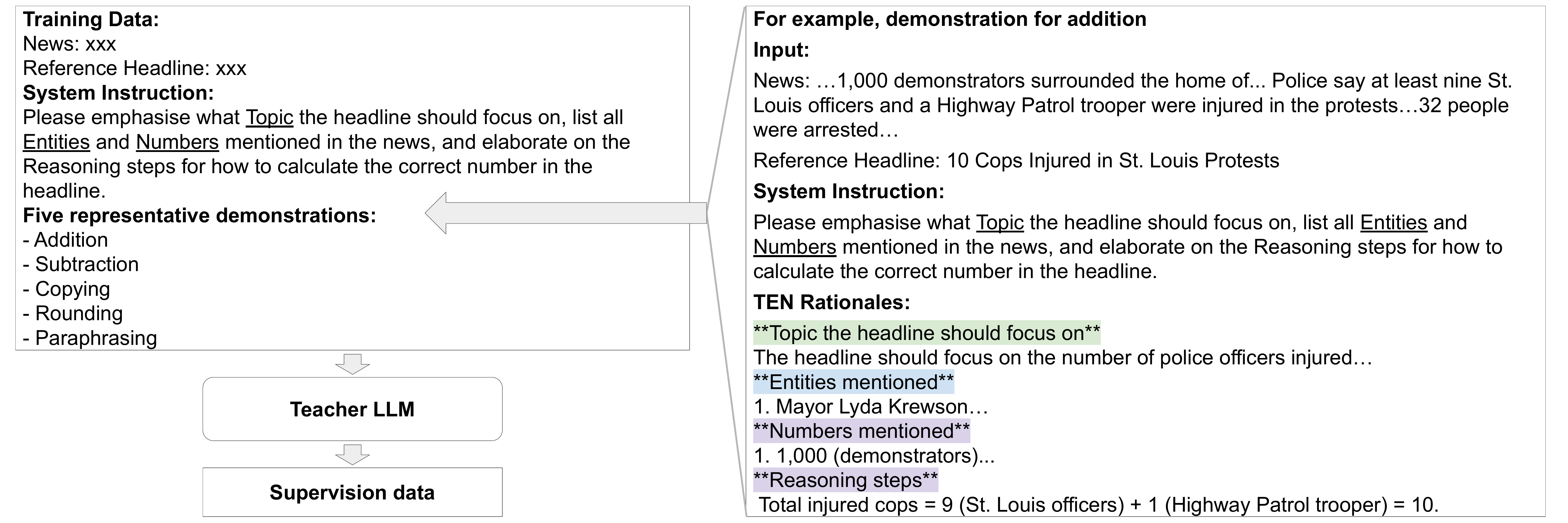}
  \caption {The process for Teacher LLM to generate TEN rationales for fine-tuning student LLMs}
  \label{fig:pre_tuning}
\end{figure*}

\subsection{Teacher LLM generation of TEN rationales}
In this phase, we focus on utilizing a teacher LLM to generate TEN rationales as supervision data. 
Figure~\ref{fig:pre_tuning} shows the process for generating this data. It is generated through a two-step process~\citep{zhang_automatic_2022}. 
In the first step, we instruct a teacher LLM with zero-shot prompting to generate demonstration TEN rationales for a few (five) representative examples for calculating numbers in headlines. 
In the second step, we employ these demonstration rationales as context to generate rationales for other examples in the whole training dataset using few-shot prompting. Specifically, we create demonstration rationales for five representative examples from the NumHG dataset \citep{huang-etal-2024-numhg}. 
The five examples are selected to represent five types of mathematical operations annotated in this dataset: (a) Copying: The numeral is directly copied from the article. (b) Addition: Numerals from the article are added to get the final numeral. (c) Subtraction: One numeral is subtracted from another. (d) Paraphrasing: The digits of the numeral are rewritten (e.g., changing 6,000 to 6k). and (e) Rounding: Only certain digits after the decimal point are retained. The details of the five demonstrations are shown in Appendix~\ref{appendix: five-demonstrations}.



We next instruct the teacher LLM to elaborate on the intermediate steps that lead to high-quality headlines with accurate numerals for the five selected examples. The prompts we use are shown in Figure~\ref{fig:pre_tuning}. 
Note that the TEN rationale comprise key elements for the news article, including numerical reasoning steps. 
Note also that to enhance the reliability of the teacher LLM-generated rationales, we also provide the reference headlines and the correct numbers to the teacher LLM in prompts. The numbers in the reference headlines are masked to ensure the teacher LLM focuses on the topic when generating topic-alignment rationales. The correct numbers are provided separately as hints to improve accuracy in numerical reasoning rationales. We then manually review and refine these five rationales to adhere to the template, ensuring consistency across the entire dataset. Using these five example TEN rationales as demonstrations, we instruct the teacher LLM to generate TEN rationales for the complete training datasets through five-shot prompting. 


\subsection{Fine-Tuning Student LLMs}
Inspired by~\citep{zhang_multimodal_2024}, we fine-tune two student LLMs independently, as illustrated in Figure~\ref{fig:three_phase_of_TEN}. The first student LLM (rationale generator) generates TEN rationales from news articles. The second model (headline generator), initialized from the same student LLM, is fine-tuned to predict headlines using both the news articles and the TEN rationales generated by the teacher LLM as inputs. 

We also apply DPO to the rationale generator to enhance its output quality. To construct the preference dataset for DPO, we first use the fine-tuned rationale generator to sample multiple rationales for each news article in the training data using a high temperature. Next, we use the headline generator to complete headlines based on the news articles and sampled rationales. We then build a pair of chosen and rejected rationales for each instance in the training data based on the following criteria: (a) Choose the rationale that leads to the headline with correct numerals and reject the one that results in a headline with incorrect numerals. (b) Choose the rationale with a high ROUGE score compared to the reference rationale and reject the one with a low ROUGE score. The flowchart for automatically constructing preference data is shown in Appendix~\ref{appendix:preference_data}. 




\section{Experiments}
We evaluate our approach TEN against state-of-the-art baselines on benchmark datasets for number-focused headline generation. 
In all experiments, GPT 4o is the teacher LLM for our TEN approach. 
All experiments were conducted on a system with 32 cores, 128GB memory and NVIDIA A100(40G) GPU. The estimated GPU usage for our experiments is approximately 2,000 hours.
All deep neural networks are implemented using Transformers~\citep{wolf2019huggingface} (distributed with the Apache-2.0 license) under the support of PyTorch~\citep{paszke2019pytorch} (distributed with the modified BSD-3-Clause license). 
Implementation details, including parameter efficient fine-tuning settings and hyperparameter settings are in Appendix~\ref{appendix:implementation}. 

\subsection{Datasets and Evaluation Metrics}
We evaluate our proposed approach using two real-world datasets: 
\begin{itemize}
	\item The NumHG dataset~\citep{huang-etal-2024-numhg} is a large dataset for number-focused headline generation that is also used for SemEval-2024 Task 7 ``NumEval: Numeral-aware language understanding and generation'' task.~\footnote{https://sites.google.com/view/numeval/numeval} 
 Each news article contains 200--300 words, and all headlines include numerals. This dataset provides human annotations of mathematical operations required to derive the numerals in each headline. 
 We apply a pre-processing step to the dataset by removing duplicate samples and retaining only those with one number in the headlines. 
 In the end, we obtained 18,315 samples for training and 3,579 for testing.
    \item The XSum dataset~\citep{narayan_dont_2018} is 
    an extreme summarization dataset comprising 226,771 BBC articles from 2010 to 2017, each accompanied by a single-sentence summary. 
    We applied pre-processing and selected articles containing 200--500 words and summaries containing only a whole number. This resulted in 9,052 samples for training and 1,605 for testing. 
\end{itemize}

We adopt the evaluation metrics commonly used in existing studies~\citep{huang-etal-2024-numhg} to assess both the textual quality and numerical accuracy for headline generation. 
We adopt ROUGE~\citep{lin2004rouge}, BERTScore~\citep{zhang2019bertscore}, and MoverScore~\citep{zhao2019moverscore} for textual quality. 
For numerical accuracy, 
a generated headline's numeral is considered correct if it matches the numeral
in the reference headline. 
We use the evaluation code~\footnote{https://github.com/ChunJiChen/NumEval\_Evaluation} from~\citet{huang-etal-2024-numhg} to automatically calculate these metrics. 


\subsection{Baselines}

We compare TEN against three baseline methods, including one representative PLM-based method and two recent LLM-based methods. 
\begin{itemize}
	\item BART~\citep{lewis-etal-2020-bart,huang-etal-2024-numhg} 
 is a PLM-based representative baseline for numerical headline generation that is shown to outperform other PLM-based methods like T5, Pegasus, SEASON, and BRIO in terms of numerical accuracy while maintaining comparable textual quality~\cite{huang-etal-2024-numhg}. 
\item NCL (NCL\_NLP)~\citep{zhao_ncl_nlp_2024} is an LLM-based method from SemEval-2024 Task 7 that achieves reasonable result. 
Similar to TEN, they also employ the teacher-student framework to generate CoT rationales to fine-tune student LLMs for headline generation. 
Different from TEN, NCL does not do the structured, element-wise rationales. 
Comparing TEN against NCL will help us understand the effectiveness of our TEN reasoning strategy. 
    \item NPP (NP-Problem)~\citep{rajpoot_team_2024} is another LLM-based baseline from SemEval-2024 Task 7. NPP achieved the highest numerical accuracy and comparable text quality among all submissions. 
They fine-tune Mistral-7B for headline generation and further align the model-generated headlines through DPO. 
As both TEN and NPP use DPO to refine headline generation, their comparison can reveal the utility of our strategy of preference data generation for DPO. 
\end{itemize}




\begin{table*}[t]
  \centering
  \resizebox{\textwidth}{!}{
  \begin{tabular}{lccc|ccc|ccc|c}
    \hline
    &&\textbf{Num Acc}&&&\textbf{ROUGE}&&&\textbf{BERTScore}&&\textbf{MoverScore}\\
    \cline{2-11}
     & Overall & Copy & Reasoning & 1 & 2 & L & P & R & F1 & \\
    \hline
    BART & 71.59 & 76.54 & 61.82 & 48.13 & 22.76 & 43.36 & 49.29 & 50.81 & 50.60 & 60.34 \\
    NPP & 74.57 & 77.43 & 68.93 & 49.24 & 23.44 & 44.08 & 50.17 & 50.57 & 50.36 & 60.54 \\
    NCL & 74.94 & 78.43 & 68.06 & 50.03 & 24.72 & 45.39 & 53.44 & 51.19 & 52.31 & \textbf{60.97} \\
    TEN (Ours) & \textbf{77.20} & \textbf{81.11} & \textbf{69.49} & \textbf{51.14} & \textbf{25.46} & \textbf{46.29} & \textbf{54.57} & \textbf{51.84} & \textbf{53.21} & \textbf{61.23} \\
    \hline
  \end{tabular}
  }
  \caption{\label{tbl:main_result_numhg}
    Numerical accuracy (\%) and textual quality score (\%) for TEN against baselines on \textbf{NumHG}. Higher numbers indicate better performance. Best results are in bold, where results within 0.5\% difference are deemed comparable.
  }
\end{table*}

\begin{table*}[t]
  \centering
  \resizebox{0.82\textwidth}{!}{
  \begin{tabular}{lc|ccc|ccc|c}
    \hline
    &\textbf{Num Acc}&&\textbf{ROUGE}&&&\textbf{BERTScore}&&\textbf{MoverScore}\\
    \cline{2-9}
     & Overall & 1 & 2 & L & P & R & F1 & \\
    \hline
    BART & 29.34 & 43.83 & 19.81 & 35.46 & 52.54 & 54.16 & 53.38 & 60.46 \\
    NPP & 30.15 & 46.32 & \textbf{22.58} & \textbf{38.08} & 55.60 & \textbf{55.82} & 55.73 & \textbf{61.35} \\
    NCL & 36.76 & 45.17 & 21.47 & 37.07 & 57.30 & 53.49 & 55.41 & \textbf{61.02} \\
    TEN (Ours) & \textbf{39.07} & \textbf{46.63} & \textbf{22.50} & \textbf{38.36} & \textbf{58.81} & 54.57 & \textbf{56.70} & \textbf{61.36} \\
    \hline
  \end{tabular}
  }
  \caption{\label{tbl:main_result_xsum}
    Numerical accuracy (\%) and textual quality metric score (\%) for TEN against baseline methods on  \textbf{XSum}. Higher numbers indicate better performance. The best results are in bold, where results within 0.5\% difference are deemed comparable.
  }
\end{table*}


\subsection{Experiment Results}

For fair comparison, all baselines use Mistral-7B, and we also use Mistral-7B as the student LLM for TEN.
Following existing studies in the literature~\cite{huang-etal-2024-numhg}, we employ numerical accuracy and textual similarity metrics ROUGE~\cite{lin2004rouge}, BERTScore~\cite{zhang2019bertscore} and MoverScore~\cite{zhao_moverscore_2019} for evaluation. 
Note that textual similarity metrics evaluate the complete headline text, including numbers as tokens, and therefore can  be seen as measuring the overall quality for number-focused headlines.

The NumHG dataset includes annotations for the type of operations needed to obtain the correct numeral in headlines. 
The numerical accuracy thus includes overall accuracy, copy accuracy (when numbers can be directly extracted from the news), and reasoning accuracy (when mathematical operations are required).   
On the XSum dataset, which lacks operation type annotations, we only report the overall accuracy. 

Observe from Table~\ref{tbl:main_result_numhg} that on the NumHG dataset, TEN achieves an overall numerical accuracy of 77.20\%, surpassing BART by 5.61\%, NPP by 2.63\%, and NCL by 2.26\% (in absolute percentage points).
In Table~\ref{tbl:main_result_xsum}, on the XSum dataset, TEN reaches an overall accuracy of 
39.07\%, outperforming BART by 9.73\%, NPP by 8.92\%, and NCL by 2.31\%. While improving numerical accuracy, TEN also maintains mostly higher textual quality by textual similarity metrics ROUGE, BERTScore, and MoverScore. 

Models trained with our TEN framework outperform NCL, demonstrating that our TEN rationales are more effective than NCL's rationales that only explain how the correct number in the headline is obtained. Our approach also outperforms NPP, demonstrating that enhancing the intermediate rationale generation process is a more effective strategy for improving headline's numerical accuracy.

We further evaluated the quality of the generated news headlines using an LLM-based metric G-Eval \cite{liu_g-eval_2023}.
Recent research shows that LLMs can be used for evaluation of quality of generated texts and demonstrate strong correlation with human judgements. 
G-Eval leverages the capability of LLMs and Chain-of-Thoughts prmopts to assess the quality of model-generated texts. 
We employed G-Eval(GPT4) to evaluate four aspects of generated headlines: coherence, consistency, fluency, and relevance. As shown in Table~\ref{tbl:llm_based_quality_evaluation}, 
TEN outperforms all baselines on NumHG and achieves comparable results on XSum.

\begin{table*}[htbp]
  \centering
  \resizebox{\textwidth}{!}{

  
  \begin{tabular}{l|cccc|cccc}
    \hline
    &\multicolumn{4}{c|}{\textbf{NumHG}} & \multicolumn{4}{c}{\textbf{XSum}}\\
    \hline    
    &\textbf{Coherence}&\textbf{Consistency}&\textbf{Fluency}&\textbf{Relevance}&\textbf{Coherence}&\textbf{Consistency}&\textbf{Fluency}&\textbf{Relevance}\\ \hline
    BART&4.0361&4.3647&2.8209&4.1068&3.3740&2.9179&2.6836&3.2840\\
    NPP&4.1734&4.6550&2.9068&4.2184&3.3782&2.9048&2.6758&3.2853\\
    NCL&4.1739&4.7108&\textbf{2.9616}&4.2264&3.3795&2.9037&2.6830&3.2767\\
    TEN&\textbf{4.1853}&\textbf{4.7210}&2.9594&\textbf{4.2436}&3.3733&2.9125&2.6802&3.2824\\
    \hline
  \end{tabular}
  }
  \caption{\label{tbl:llm_based_quality_evaluation}
    G-Eval scores for TEN against baselines on \textbf{NumHG} and \textbf{XSum}. Headlines are assessed in terms of Coherence (1-5), Consistency (1-5), Fluency (1-3), and Relevance (1-5). The higher numbers indicate better performance. Best results are in bold.}
\end{table*}

\subsection{Performance of TEN rationales and Teacher-student knowledge distillation}
The structured rationales and teacher-student paradigm to distill knowledge from a teacher LLM (GPT-4o) to a student LLM (Mistral 7B and Llama 3.1) are important parts of our TEN framework. 
To evaluate the effectiveness of TEN rationales and if GPT-4o is a good teacher LLM, we evaluated the performance of GPT-4o with and without TEN structured rationales under zero-shot prompting. 
The results are shown in Table~\ref{tbl:main_result_gpt4o_both}. It can be seen that by prompting GPT-4o to generate TEN rationales, performance improved significantly both for numerical accuracy and textual quality.

\begin{table*}[htbp]
  \centering
  \resizebox{\textwidth}{!}{
  \begin{tabular}{l|cccc|cccc}
    \hline
    &\multicolumn{4}{c|}{\textbf{NumHG}} & \multicolumn{4}{c}{\textbf{XSum}}\\
    \hline
     & Num Acc & ROUGE-1 & BERTScore-F1 & MoverScore & Numm Acc & ROUGE-1 & BERTScore-F1 & MoverScore\\
    \hline
    w/o TEN & 21.39 & 35.59 & 39.21 & 56.53 & 6.23 & 22.59 & 22.76 & 54.58\\
    w TEN & 33.01 & 35.47 & 39.86 & 56.72 & 9.91 & 22.18 & 23.14 & 54.63\\
    \hline
  \end{tabular}
  
  }
\caption{\label{tbl:main_result_gpt4o_both}
Performance of GPT-4o with/without TEN rationales under zero-shot prompting on \textbf{NumHG} and \textbf{XSum}.
  }
\end{table*}

We further conducted experiments to evaluate if the student LLM can effectively learn rationale generation from the teacher LLM. 
On the test data, we computed the textual and semantic similarity scores for the rationales automatically generated by Mistral-7B-v0.3 and Llama-3.1-8B as the student model, against the supervision data generated by the teacher LLM GPT-4o. Table~\ref{tbl:rationale_evaluation} illustrates the evaluation results. The high textual similarity and semantic similarity scores demonstrate that the student model can learn from the teacher model to generate rationales to enhance its reasoning capability for number-focused headline generation.

\begin{table*}[htbp]
  \centering
  \resizebox{\textwidth}{!}{
  \begin{tabular}{l|ccc|ccc}
    \hline
    &\multicolumn{3}{c|}{\textbf{NumHG}} & \multicolumn{3}{c}{\textbf{XSum}}\\
    \hline
     & ROUGE-1 & BERTScore-F1 & MoverScore & ROUGE-1 & BERTScore-F1 & MoverScore\\
    \hline
    Mistral-7B-v0.3 & 84.12 & 81.13 & 71.03 & 75.46 & 70.43 & 66.68\\
    Llama-3.1-8B & 84.02 & 80.90 & 70.91 & 75.35 & 70.18 & 66.69\\
    \hline
  \end{tabular}
  }
  \caption{\label{tbl:rationale_evaluation}
    textual quality metric score (\%) for Student LLM generated rationales against teache LLM generated supervision rationales on \textbf{NumHG} and \textbf{XSum}.
  }
\end{table*}

\subsection{Ablation study}

\noindent \textbf{Effect of refining rationales through DPO.} In TEN, we apply DPO to enhance the capability of the student LLM rationale generator for topic alignment and numerical reasoning. 
To understand the effectiveness of DPO we tested TEN minus($-$) DPO on both Mistral-7B and Llama-3.1-8B. Table~\ref{tbl:ablation} illustrates the results. It can be seen that using a rationale generator improved through DPO leads to higher numerical accuracy and textual quality. On the NumHG dataset, DPO improves the numerical accuracy by 0.96\% with a Mistral-7B-v0.3-based rationale generator, and by 1.83\% with Llama-3.1-8B. On the XSum dataset, DPO enhances the numerical accuracy of Mistral-7B-v0.3 and Llama-3.1-8B by 1.38\% and 0.75\%, respectively. Additionally, DPO enhances the ROUGE scores marginally for both student models across both benchmark datasets. 

\begin{table*}[t]
  \centering
  \resizebox{\textwidth}{!}{
  \begin{tabular}{lcccccccc}
    \hline
    \textbf{Method} & \multicolumn{4}{c}{\textbf{NumHG}} & \multicolumn{4}{c}{\textbf{XSum}} \\
    \hline
    & NumAcc & ROUGE-1 & BERTScore-F1 & MoverScore & NumAcc & ROUGE-1 & BERTScore-F1 & MoverScore \\
    \hline
    Mistral-7B-v0.3 & & & & & & & \\
    \hline
    \: TEN (Ours) & \textbf{77.20} & \textbf{51.14} & \textbf{53.21} & \textbf{61.23} & \textbf{39.07} & 46.63 & \textbf{56.70} & \textbf{61.36} \\
    \: $-$ DPO & 76.24 & 50.95 & 53.12 & \textbf{61.24} & 37.69 & 46.26 & 56.35 & \textbf{61.27}  \\
    \: $-$ DPO $-$ N & 74.04 & 50.23 & 52.59 & \textbf{61.08} & 35.58 & 46.18 & 56.13 & \textbf{61.21}  \\
    \: $-$ DPO $-$ TE & 75.55 & \textbf{51.63} & \textbf{53.67} & \textbf{61.37} & 31.71 & 45.94 & 55.55 & \textbf{61.11}  \\
    \: $-$ DPO $-$ TEN & 70.33 & \textbf{51.27} & \textbf{53.43} & \textbf{61.42} & 30.41 & \textbf{47.37} & \textbf{56.88} & \textbf{61.58} \\
    \hline
    \hline
    Llama-3.1-8B & & & & & & & &  \\
    \hline
    \: TEN (Ours) & \textbf{77.89} & \textbf{51.15} & \textbf{52.83} & \textbf{61.14} & \textbf{37.51} & 46.11 & \textbf{56.09} & \textbf{61.18} \\
    \: $-$ DPO & 76.06 & \textbf{50.69} & 52.59 & \textbf{61.09} & 36.76 & 45.89 & 55.80 & \textbf{61.06}  \\
    \: $-$ DPO $-$ N & 73.80 & 50.05 & 52.24 & \textbf{60.90} & 36.51 & 45.83 & \textbf{56.03} & \textbf{61.11} \\
    \: $-$ DPO $-$ TE  & 74.86 & \textbf{50.98} & \textbf{53.21} & \textbf{61.22} & 32.02 & 45.62 & 55.50 & \textbf{61.03}  \\
    \: $-$ DPO $-$ TEN & 70.71 & \textbf{51.01} & \textbf{53.21} & \textbf{61.34} & 29.35 & \textbf{46.62} & \textbf{56.44} & \textbf{61.43} \\    
    \hline
  \end{tabular}
  }
  \caption{\label{tbl:ablation}
    Results (\%) for ablation study of TEN. Higher numbers indicate better performance. Best results are in bold, where results within 0.5 (\%) difference are deemed comparable.     
  }
\end{table*}



\noindent \textbf{Effect of different supervision signals}. In TEN, we have developed two types of CoT supervision signals. One focuses on aligning the topic of the generated headline, while the other enhances numerical calculation accuracy. As illustrated in Figure~\ref{fig:three_phase_of_TEN}, the rationales under "Topic" and "Entities" contribute to topic alignment, whereas those under "Numbers mentioned" and "Reasoning steps" boost numerical reasoning. We've assessed the impact of these supervision signals, with results for ``TEN minus Number'' ($-$ N), ``TEN minus Topic and Entity ''($-$ TE), and ''no supervision''($-$ TEN) presented in Table~\ref{tbl:ablation}. It can be seen that both types of signals independently improve numerical accuracy in headline generation. However, their effectiveness varies across the two benchmark datasets: numerical reasoning signals show a more pronounced effect on NumHG, while topic alignment signals have a greater impact on XSum. Notably, combining both types of supervision signals yields optimal model performance with highest numerical accuracy and comparable textual quality. 



\noindent \textbf{TEN performance with different student LLMs.} We also want to highlight the performance of TEN with different base student LLMs. All results in Table~\ref{tbl:ablation} are obtained using GPT-4o as the teacher LLM and two different student LLMs: Mistral-7B and Llama-3.1-8B. 
Observe that Mistral-7B and Llama-3.1 demonstrate similar performance for both numerical accuracy and textual quality. 
It can also be seen that the supervision signals and DPO show their effectiveness for both student LLMs. 


\begin{figure*}[t!]
  \includegraphics[width=0.48\linewidth]{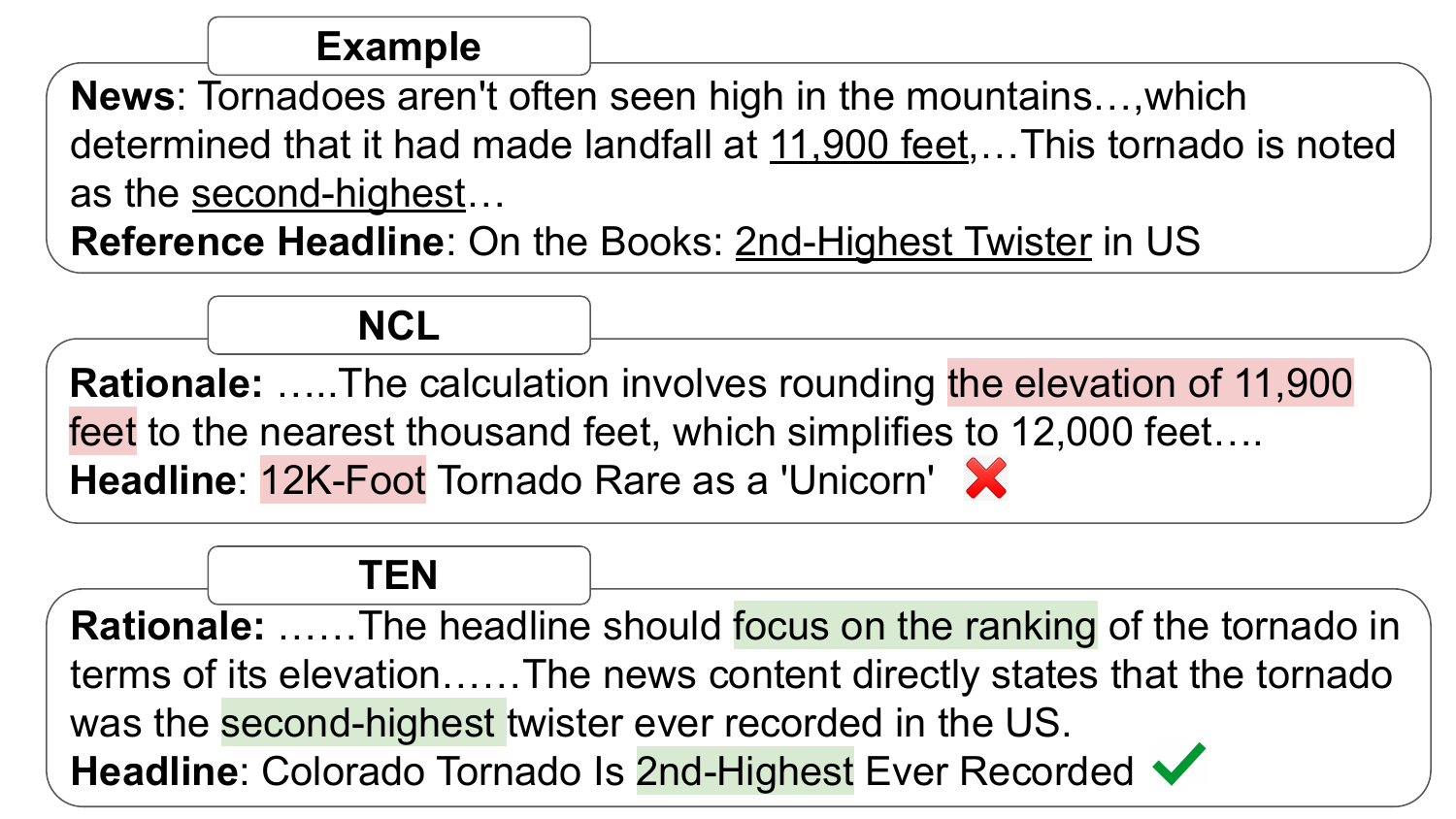} \hfill
  \includegraphics[width=0.48\linewidth]{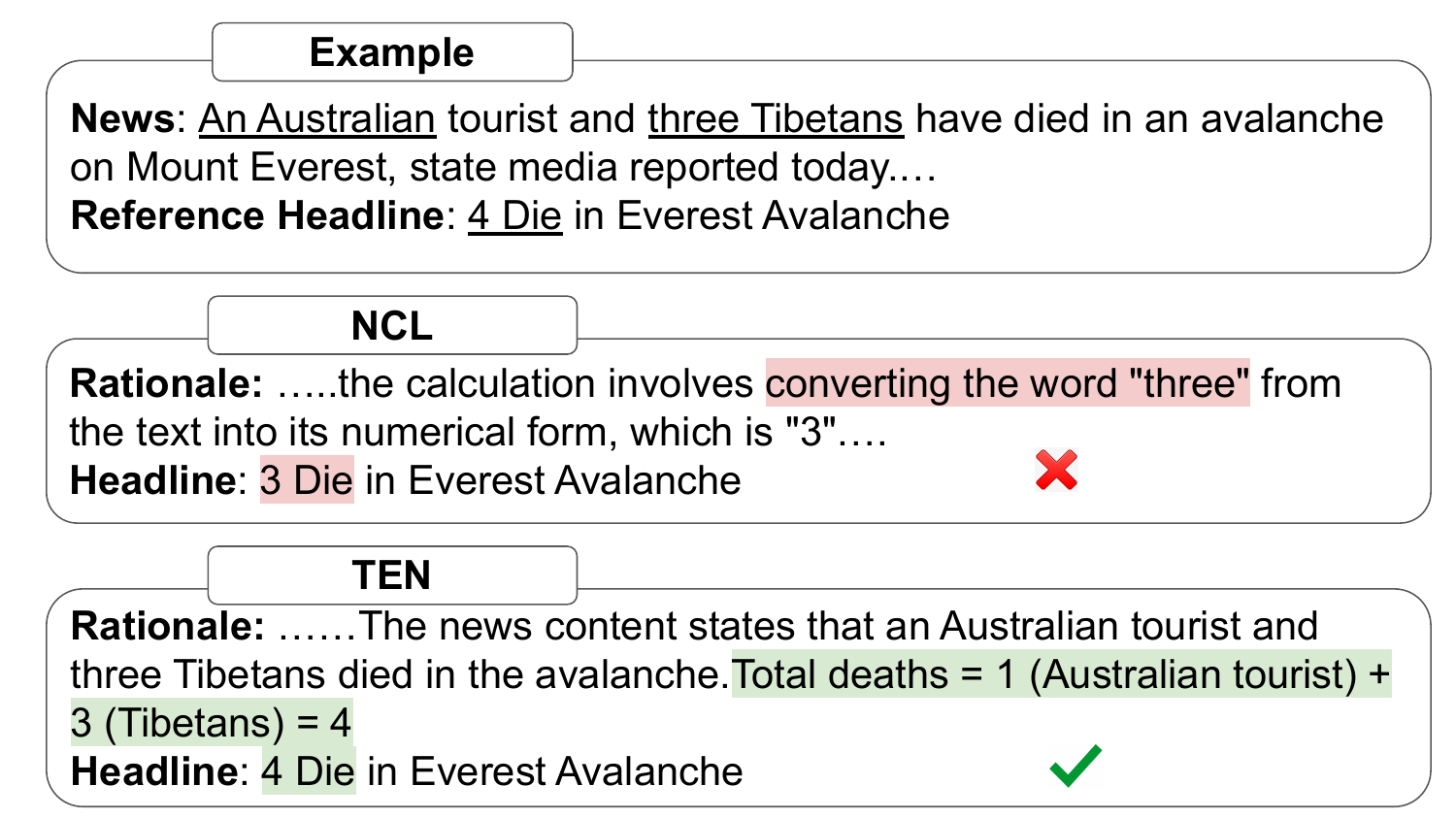}
  \parbox[b]{0.45\textwidth}{\centering (a) Topic alignment}
  \hspace{0.05\textwidth}
  \parbox[b]{0.45\textwidth}{\centering (b) Numerical reasoning }  
  \caption {\label{fig:case_study} TEN vs. NCL (Baseline) for rationale and headline generation}
\end{figure*}

\subsection{Error analysis and case study}
\begin{table*}[!h]
    \centering
    \resizebox{0.9\textwidth}{!}{
    \begin{tabular}{ccccccccccc}
    \hline
    Operation & Overall & Copy & Trans & Paraphrase & Round & Subtract & Add & Span & Divide & Multiply \\
    Count & 3996 & 2,494 & 682 & 376 & 133 & 89 & 76 & 85 & 28 & 33 \\
    \hline
    BART (Err\%) & 31.53 & 23.46 & 34.02 & 26.60 & 60.90 & 96.63 & 78.95 & 68.24 & 96.43 & 93.94 \\
    NPP (Err\%) & 27.55 & 22.57 & 30.06 & 21.81 & \textbf{40.60} & \textbf{68.54} & \textbf{56.58} & \textbf{49.41} & \textbf{82.14} & 84.85 \\
    NCL (Err\%) & 28.08 & 21.57 & 29.62 & 20.74 & 48.12 & 79.78 & 75.00 & 64.71 & 96.43 & 90.91 \\
    TEN (Err\%) & \textbf{25.40} & \textbf{18.89} & \textbf{27.57} & \textbf{20.21} & 48.12 & 75.28 & 60.53 & 58.82 & 92.86 & \textbf{81.82} \\
    \hline
    \end{tabular}
}
  \caption{\label{tbl:error_analysis}
    Error analysis across different mathematical operations on test data from NumHG. Lower numbers indicate better performance. Best results are in bold, where results within 0.5\% difference are deemed comparable. 
  }
\end{table*}

\noindent \textbf{Error analysis.} Utilizing the annotations from the NumHG dataset, which outlines nine types of operations necessary for calculating numerals in headlines, we conducted an error analysis for TEN in comparison to the baselines. We present the error rates in Table~\ref{tbl:error_analysis}. 
The experimental results demonstrate that our approach significantly reduces errors in copying, translating, and paraphrasing, achieving the lowest error rates compared to baseline methods. These three operations represent over 88\% of the total. For the remaining less frequent operations, our approach achieves error rates comparable to the best-performing baseline.

\noindent \textbf{Case study.} Two examples are selected from the test dataset to 
illustrate the benefits of the TEN reasoning strategy, compared against the NCL baseline, which generates rationale without the TEN structured rationales. 
Figure~\ref{fig:case_study} (a) shows that TEN correctly identifies the topic the headline should focus on in the rationale, which is the rank of the tornado in this case, while NCL mistakenly focuses on the elevation. In Figure~\ref{fig:case_study} (b), TEN successfully calculates the number of people who died by adding 1 Australian tourist and 3 Tibetans, while NCL fails to count the Australian tourist. 

\section{Conclusion}
In this paper
we studied number-focused news headline generation, 
a problem presenting the unique challenge of high textual quality with precise number accuracy for LLM generation. 
We proposed a novel framework of using rationales of key elements Topic, Entity, and Numerical reasoning (TEN) to enhance the capability of LLMs for topic alignment and numerical reasoning in headline generation.   
We developed an approach to fine-tune LLMs to automatically generate TEN rationales for numerical headlines generation. 
Especially our TEN approach builds upon the teacher-student rationale-augmented training framework, where a teacher LLM automatically generate TEN rationales as supervision data to teach a student LLM rationale generator and a student LLM headline generator.    
Experiments on popular numerical news headline generation datasets showed that TEN outperforms existing approaches, achieving higher numerical accuracy and mostly better textual quality for headline generation. 

\section*{Acknowledgements}
This research is supported in part by the Australian Research Council Discovery Project DP200101441.

\section*{Limitations}

One limitation of our study is that due to computing resource limitation, we have only applied parameter-efficient technique QLoRA \citep{dettmers_qlora_2023} to fine-tune student LLMs, 
and as such
it is possible that we 
have not fully elicited the capability of student LLMs. 
Another limitation of our study is the limited data for experiments. 
To our best knowledge NumHG is the only public benchmark dataset for numerical headline generation, and we constructed one more dataset based on XSum for extreme summarization.  


\clearpage

\onecolumn
\appendix

\section{Flowchart for Constructing Preference Data}
\label{appendix:preference_data}
\begin{figure}[!h]
  \centering
  \includegraphics[width=\textwidth]{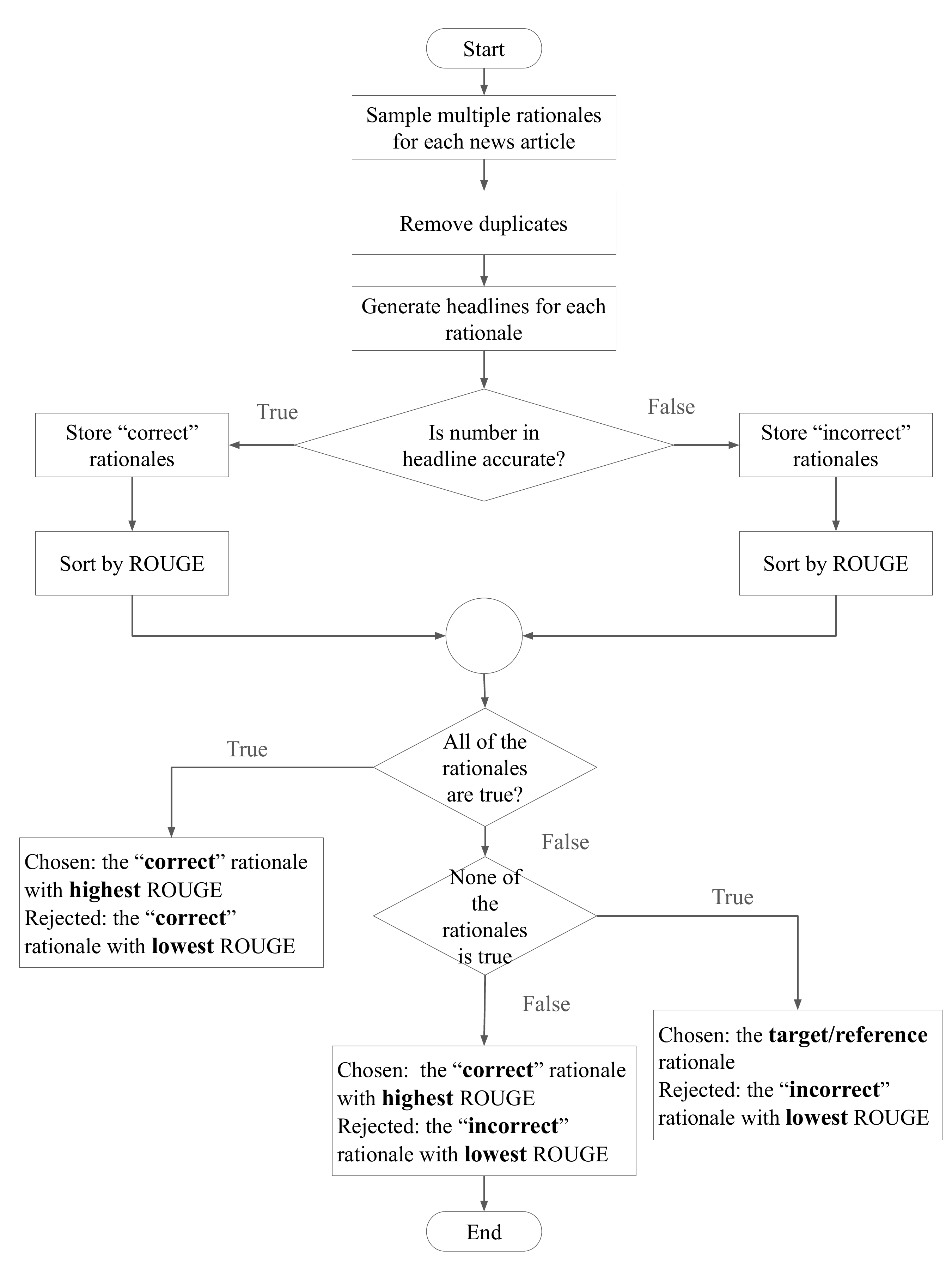}
  \caption{Flowchart for constructing preference data}
  \label{appendix:constructing_preference_data}
\end{figure}

Figure~\ref{appendix:constructing_preference_data} shows the flowchart for the construction of a pair of chosen-rejected rationales for a given news article in the training data. We apply this process and construct a dataset containing preferences for all the instance in the training data. Then the preference data is used for DPO. 


\section{Implementation Details}
\label{appendix:implementation}
We employ GPT-4o as the teacher LLM to generate supervision data and experiment with two student LLMs: Mistral-7B-v0.3 and Llama-3.1-8B. All our rationale generators and headline generators are fine-tuned on the training data for three epochs. We apply QLoRA \citep{dettmers_qlora_2023} techniques to fine-tune the student LLMs efficiently. For all rationale generators, we set the LoRA rank and LoRA alpha to 128 and 64, respectively. For all headline generators, we set these values to 64 and 32. We fine-tune Mistral-7B-v0.3 with a learning rate of 2e-4 and Llama-3.1-8B with 8e-4. To create the preference dataset of chosen and rejected rationales, we fine-tune Mistral-7B-v0.3 and Llama-3.1-8B on the training data for a single epoch, sample 15 TEN rationales from each model for every news article, remove duplicate generations, and create a pair of chosen and rejected rationales for each training sample. During DPO, we set the LoRA rank to 256, LoRA alpha to 128, and DPO beta to 0.8. We set the learning rates for Mistral-7B-v0.3 and Llama-3.1-8B to 2e-6 and 8e-6 when doing experiments on NumHG, while setting them to 5e-6 and 2e-5 on XSum.


\clearpage
\section{Generation of Supervision Data}
\label{appendix: five-demonstrations}

The following five examples are demonstrations for five distinct operations needed to calculate the correct numbers in the headlines. The outputs (TEN rationales) are generated by the GPT-4o via zero-shot prompting respectively. Then the five examples, including generated TEN rationales, are used as demonstrations for the GPT-4o to generate supervision data for the entire training dataset.

\begin{table}[htbp]
    \centering
    \begin{tabular}{p{0.9\textwidth}}
    \hline
    \textbf{Example 1: Copy} \\
    \hline
    \textbf{Input:} \\
    \textbf{News:} (Oct 29, 2013 8:15 AM CDT) Dax Shepard and Kristen Bell got married at the Beverly Hills courthouse, in a ceremony about as different from Kim Kardashian's last wedding extravaganza as it is possible to be. As Shepard revealed last night on Jimmy Kimmel Live, the whole thing—including the fuel it took to get to the courthouse—cost \$142. It was just Kristen and I at this lonely courthouse, he said, so friends showed up afterward with a cake reading, in icing, The World's Worst Wedding. How many people can say they threw the world's worst wedding? Shepard asked. \\
    \textbf{Headline:} Dax Shepard: Wedding to Kristen Bell Cost \$\_\_\_\_. \\
    \textbf{Correct Number:} 142 \\
    \textbf{System Instruction: } You will be given a piece of news after prefix "News: ". You will be given a masked headline of the news after prefix "Headline:". You will be provided the correct missing numeral after prefix "Correct Number:". Please emphasise what Topic the headline should focus on, list all Entities and Numbers mentioned in the news, and elaborate on the Reasoning steps for how to calculate the correct number in the headline. \\
    \hline
    \textbf{Output:} \\
    \***Topic the headline should focus on** \\
    The headline should focus on the cost of Dax Shepard and Kristen Bell's wedding. \\
    \\
    \***Entities Mentioned** \\
    1. Dax Shepard \\
    2. Kristen Bell \\
    3. Kim Kardashian \\
    4. Jimmy Kimmel \\
    \\
    \***Numbers Mentioned** \\
    1. \$142 (cost of the wedding) \\
    \\
    \***Reasoning steps** \\
    The news content directly states that the entire wedding, including the fuel to get to the courthouse, cost \$142. \\
    
    Therefore, the numeral in the headline should be 142. \\
    \hline
    \end{tabular}
  
\end{table}

\begin{table}[htbp]
    \centering
    \begin{tabular}{p{0.9\textwidth}}
    \hline
    \textbf{Example 2: Add} \\
    \hline
    \textbf{Input:} \\
    \textbf{News:} (Sep 16, 2017 8:26 AM CDT) Police in St. Louis used tear gas to break up protests Friday night after around 1,000 demonstrators surrounded the home of Mayor Lyda Krewson, hurling red paint and breaking a front window. Police say at least nine St. Louis officers and a Highway Patrol trooper were injured in the protests, which followed the acquittal of white former police officer Jason Stockley in the 2011 shooting death of black motorist Anthony Lamar Smith, the St. Louis Post-Dispatch reports. Police say 32 people were arrested when a demonstration that had been largely peaceful turned violent after dark and agitators began to destroy property and assault police officers. Activists have promised more protests against the acquittal of Stockley, who, during a police chase, was caught on dashcam video saying I'm going to kill this motherf*****, don't you know it, the AP reports. Less than a minute later, he shot Smith five times.....  \\
    \textbf{Headline:} \_\_\_\_ Cops Injured in St. Louis Protests. \\
    \textbf{Correct Number:} 10 \\
    \textbf{System Instruction: } You will be given a piece of news after prefix "News: ". You will be given a masked headline of the news after prefix "Headline:". You will be provided the correct missing numeral after prefix "Correct Number:". Please emphasise what Topic the headline should focus on, list all Entities and Numbers mentioned in the news, and elaborate on the Reasoning steps for how to calculate the correct number in the headline. \\
    \hline
    \textbf{Output:} \\
    \***Topic the headline should focus on** \\
    The headline should focus on the number of police officers injured during the St. Louis protests. \\
    \\
    \***Entities Mentioned:**\\
    1. Mayor Lyda Krewson \\
    2. Jason Stockley (white former police officer) \\
    3. Anthony Lamar Smith (black motorist) \\
    4. Reverend Clinton Stancil \\
    5. St. Louis officers \\
    6. Highway Patrol trooper \\
    \\
    \***Numbers Mentioned** \\
    1. 1,000 (demonstrators) \\
    2. 9 (St. Louis officers injured) \\
    3. 1 (Highway Patrol trooper injured) \\
    4. 2011 (year of the shooting) \\
    5. 32 (people arrested) \\
    6. 3 (years since Ferguson unrest) \\
    7. 18 (age of Michael Brown) \\
    8. 5 (times Smith was shot) \\
    9. 24 (age of Anthony Lamar Smith) \\
    \\
    \***Reasoning steps** \\
    To find the total number of cops injured in the St. Louis protests, we need to sum the number of St. Louis officers injured and the Highway Patrol trooper injured. \\
    
    - St. Louis officers injured: 9 \\
    - Highway Patrol trooper injured: 1 \\

    Total injured cops = 9 (St. Louis officers) + 1 (Highway Patrol trooper) = 10 \\
    
    Therefore, the numeral in the headline should be 10. \\
    \hline
    \end{tabular}
  
\end{table}

\begin{table}[htbp]
    \centering
    \begin{tabular}{p{0.9\textwidth}}
    \hline
    \textbf{Example 3: Subtract} \\
    \hline
    \textbf{Input:} \\
    \textbf{News:} (Jun 26, 2017 9:30 AM CDT) After three decades on the lam, an Arkansas man has been quickly reacquainted with life behind bars. Steven Dishman, 60—who was five months into a seven-year sentence for burglary and property theft when he escaped from the Cummins Unit in Lincoln County, Ark., on May 28, 1985, reports KTHV—was arrested Sunday at a home in Springdale in the northwestern part of the state, report Arkansas Online and KFSM. Arkansas State Police and local cops raided the home after receiving a tip. A rep for the Arkansas Department of Correction tells Fox 16 that Dishman will now be required to serve the remainder of his sentence. Dishman may also face additional charges related to his escape, the details of which were not immediately explained. Had Dishman stayed in jail, he would have been eligible for parole in 1987 and freed by 1991.\\
    \textbf{Headline:} He Spent \_\_\_\_ Years on the Run. Then Police Received a Tip. \\
    \textbf{Correct Number:} 32 \\
    \textbf{System Instruction: } You will be given a piece of news after prefix "News: ". You will be given a masked headline of the news after prefix "Headline:". You will be provided the correct missing numeral after prefix "Correct Number:". Please emphasise what Topic the headline should focus on, list all Entities and Numbers mentioned in the news, and elaborate on the Reasoning steps for how to calculate the correct number in the headline. \\
    \hline
    \textbf{Output:} \\
    \***Topic the headline should focus on** \\
    The headline should focus on the number of years Steven Dishman spent on the run before being recaptured. \\
    \\
    \***Entities Mentioned** \\
    1. Steven Dishman \\
    2. Arkansas State Police \\
    3. Local cops \\
    4. Arkansas Department of Correction representative \\
    \\
    \***Numbers Mentioned** \\
    1. 60 (age of Steven Dishman) \\
    2. 7 (years sentence for burglary and property theft) \\
    3. 5 (months into his sentence when he escaped) \\
    4. May 28, 1985 (date of escape) \\
    5. 1987 (year he would have been eligible for parole) \\
    6. 1991 (year he would have been freed) \\
    7. 2017 (current year) \\
     \\
    \***Reasoning steps** \\
    To find the number of years Steven Dishman spent on the run, we need to calculate the difference between the year he escaped (1985) and the year he was recaptured (2017). \\
    
    - Year of escape: 1985 \\
    - Year of recapture: 2017 \\
    
    Years on the run = 2017 - 1985 = 32 \\
    
    Therefore, the numeral in the headline should be 32. \\
    \hline
    \end{tabular}
  
\end{table}

\begin{table}[htbp]
    \centering
    \begin{tabular}{p{0.9\textwidth}}
    \hline
    \textbf{Example 4: Paraphrase} \\
    \hline
    \textbf{Input:} \\
    \textbf{News:} News: (Oct 7, 2014 12:40 PM CDT) As of Jan. 1, Walmart will no longer offer 30,000 of its employees health insurance. Bloomberg notes that's about 2\% of its workforce. The move comes as a reaction to the company's rising health care costs as far more of its employees and their families enrolled in its health care plans than it had expected following the ObamaCare rollout. The AP reports those costs will surge \$500 million this fiscal year, \$170 million more than had been estimated. Those affected are employees who average fewer than 30 hours of work per week; the Wall Street Journal explains they were grandfathered in when Walmart in 2012 stopped offering insurance to new hires who didn't exceed the 30-hour threshold. A benefits expert says Walmart is actually late to the game in terms of cutting insurance to some part-time workers; Target, the Home Depot, and others have already done so. Meanwhile, Walmart's full time workers will see their premiums rise in 2015. Premiums for the basic plan, which 40\% of its workforce is on, will increase 19\% to \$21.90 per pay period come Jan. 1. \\
    \textbf{Headline:} Headline: \_\_\_\_K Walmart Part-Timers to Lose Health Insurance. \\
    \textbf{Correct Number:} 30. \\
    \textbf{System Instruction: } You will be given a piece of news after prefix "News: ". You will be given a masked headline of the news after prefix "Headline:". You will be provided the correct missing numeral after prefix "Correct Number:". Please emphasise what Topic the headline should focus on, list all Entities and Numbers mentioned in the news, and elaborate on the Reasoning steps for how to calculate the correct number in the headline. \\
    \hline
    \textbf{Output:} \\
    \***Topic the headline should focus on** \\
    The headline should focus on the number of Walmart part-time employees who will lose their health insurance. \\
    \\
    \***Entitiess Mentioned** \\
    1. Walmart employees \\
    2. Walmart's full-time workers \\
    3. Target \\
    4. The Home Depot \\
    5. ObamaCare \\
    \\
    \***Numbers Mentioned** \\
    1. 30,000 (employees losing health insurance) \\
    2. 2\% (of Walmart's workforce) \\
    3. \$500 million (surge in health care costs) \\
    4. \$170 million (more than estimated) \\
    5. 30 hours (work per week threshold) \\
    6. 2012 (year when Walmart stopped offering insurance to new hires under 30 hours) \\
    7. 2015 (year when premiums will rise) \\
    8. 40\% (of workforce on the basic plan) \\
    9. 19\% (increase in premiums) \\
    10. \$21.90 (new premium per pay period) \\
    \\
    \***Reasoning steps** \\
    The news content directly states that Walmart will no longer offer health insurance to 30,000 of its employees. \\

    To convert this number into thousands (K): \\
    - 30,000 employees = 30K employees \\
    
    Therefore, the numeral in the headline should be 30. \\    \hline
    \end{tabular}
  
\end{table}

\begin{table}[htbp]
    \centering
    \begin{tabular}{p{0.9\textwidth}}
    \hline
    \textbf{Example 5: Rounding} \\
    \hline
    \textbf{Input:} \\
    \textbf{News:} (Apr 6, 2011 10:10 AM CDT) Well that was fast. Scott Walker did an about-face yesterday and demoted the son of a major campaign donor from his \$81,500-per-year Commerce Department gig—to his former \$64,728 job. Brian Deschane will return to his previous post as a bureau director at the Department of Regulation and Licensing, the Milwaukee Journal-Sentinel reports. The 27-year-old, who has no college degree and little experience (but does have two drunk-driving convictions!), had been in that job just two months when he was promoted to head of Wisconsin's environmental and regulatory affairs. Among those Deschane beat out for the job: a former Cabinet secretary with a doctoral degree and eight years experience overseeing oil contamination cleanup, and a professional engineer who had held the job since 2003 under the last governor. Neither was even interviewed. When Governor Walker learned the details of this agency staffing decision, he directed his administration to move in another direction, a Walker spokesman says (last week, this same rep called Deschane a natural fit for the position). But the demotion didn't satisfy some Democrats. I want to make sure he was properly hired for that job in the first place, said Assembly Minority Leader Peter Barca. Click for more on the brouhaha. \\
    \textbf{Headline:}  Headline: Walker Demotes Donor's Kid— to Puny \$\_\_\_\_K Job. \\
    \textbf{Correct Number:} 65. \\
    \textbf{System Instruction: } You will be given a piece of news after prefix "News: ". You will be given a masked headline of the news after prefix "Headline:". You will be provided the correct missing numeral after prefix "Correct Number:". Please emphasise what Topic the headline should focus on, list all Entities and Numbers mentioned in the news, and elaborate on the Reasoning steps for how to calculate the correct number in the headline. \\
    \hline
    \textbf{Output:} \\
    \***Topic the headline should focus on** \\
    The headline should focus on the new, lower salary of the demoted son of a major campaign donor. \\
    \\
    \***Entities Mentioned** \\
    1. Scott Walker \\
    2. Brian Deschane \\
    3. Assembly Minority Leader Peter Barca \\
    4. Former Cabinet secretary (unnamed) \\
    5. Professional engineer (unnamed) \\
    \\
    \***Numbers Mentioned** \\
    1. \$81,500 (initial salary at Commerce Department) \\
    2. \$64,728 (new salary after demotion) \\
    3. 27 (age of Brian Deschane) \\
    4. 2 (months in the initial job before promotion) \\
    5. 2003 (year since the professional engineer held the job) \\
    6. 8 (years of experience of the former Cabinet secretary) \\
    \\
    \***Reasoning Steps** \\
    The news content states that Brian Deschane was demoted from his \$81,500-per-year job to his former \$64,728 job. \\
    
    To convert this new salary into thousands (K): \\
    $\$64,728 \approx \$65,000$ \\
    
    Therefore, the numeral in the headline should be 65. \\
    \hline
    \end{tabular}
  
\end{table}

\end{document}